\documentclass[runningheads,a4paper]{llncs}

\usepackage{amssymb}
\usepackage{graphicx}
\graphicspath{ {./images/} }
\usepackage{xcolor}
\usepackage{bibnames}

\usepackage{listings}
\usepackage{url}
\urldef{\mailsa}\path||
\urldef{\mailsb}\path||
\urldef{\mailsc}\path||    
\newcommand{\keywords}[1]{\par\addvspace\baselineskip
\noindent\keywordname\enspace\ignorespaces#1}

\begin{document}

\mainmatter  

\title{ROS Rescue :  Fault Tolerance System for Robot Operating System}

\titlerunning{ROS Rescue :  Fault Tolerance System for Robot Operating System}

%
%
\author{Pushyami Kaveti, Hanumant Singh}
\authorrunning{Pushyami Kaveti, Hanumant Singh}

\institute{
Northeastern University, Boston, MA 02130\\
\mailsa kaveti.p@husky.neu.edu\\
\mailsb ha.singh@northeastern.edu\\
\mailsc\\
\url{https://www.northeastern.edu/robotics}}

%
%

\toctitle{Lecture Notes in Computer Science}
\tocauthor{Authors' Instructions}
\maketitle

\begin{abstract}
In this chapter we discuss the problem of master failure in ROS1.0 and its impact on robotic deployments in the real world. We address this issue in this tutorial chapter where we outline, design and demonstrate a fault tolerant mechanism associated with ROS master failure. Unlike previous solutions which use primary backup replication and external checkpointing libraries which are process heavy, our mechanism adds a lightweight  functionality to the ROS master to enable it to recover from failure.

We present a modified version of ROS master which is equipped with a logging mechanism to record the meta information and network state of ROS nodes as well as a recovery mechanism to go back to the previous state without having to abort or restart all the nodes. We also implement an additional master monitor node responsible for failure detection on the master by polling it for its availability. Our code is implemented in python and preliminary tests were conducted successfully on a variety of land, aerial and underwater robots and a tele-operating computer running ROS Kinetic on Ubuntu 16.04. The code is publicly available under a creative commons license  on github at \url{https://github.com/PushyamiKaveti/fault-tolerant-ros-master}.

\end{abstract}

\section{Introduction}
The Robot Operating System (ROS)\cite{Quigley2009} is the most widely used framework \cite{ros} that provides libraries and tools to create robotic applications. It allows us to build and execute distributed applications across multiple machines within the context of a publish/subscribe master-slave architecture. ROS supports both synchronous communication via services and  asynchronous communication via publish/subscribe. It maintains a peer to peer runtime graph of processes communicating via XMLRPC.

\par The ROS master has a unique place in the ROS architecture. It acts as a central authority for the nodes to setup communication links with each other. This uniqueness is a major issue as the system is prone to a single point of failure when the ROS master crashes. Even though a ROS master crashing may be a very low probability event, a requirement for resiliency to faults is critical for implementing real world robotics applications.
\par We have experienced ROS master crashes in the environments that we focus on - long deployments on missions in challenging environments with extremes of temperature and adverse environmental conditions as are found in underwater robotic applications and in  ice/snow covered areas. Battery limitations in case of high power consumption on Unmanned Aerial Systems (UAS or drones), software failures associated with memory or processing overload can also be contributing factors. The impact of failure may be dramatic and catastrophic leading to the loss of the robot, property damage and possible human injury or death when dealing with mission critical systems on large robotic systems such as unmanned aerial systems and autonomous cars. Thus, it is extremely important to detect and recover from such failures, and to do so quickly and without the need to abort or reboot the mission entirely.

\subsection{Related work}  
 There have been previous attempts towards fault tolerance in ROS \cite{lauer2018resilient}\cite{jain2017dmtcp}. Both these cases propose a transparent approach where there is a separation between functional code and the fault tolerance mechanism.

\par Lauer et al \cite{lauer2018resilient}, propose an adaptive fault-tolerance technique related to a component based approach. Their design to fault-tolerance uses primary-backup replication and the introduction of, \textit{before-proceed-after}, interceptor nodes between the clients (ROS nodes) and the master. A back-up replica of the ROS master is maintained to tolerate faults associated with crashes. A request to the master is first handled by a proxy server which is then passed on to the \textit{before-proceed-after} nodes. Once it passes through the \textit{after} node, the backup replica is updated to match the primary. There is a crash detector node on the master which periodically reports master’s status to the crash detector on the slave. There is also a recovery node which is notified in the event of a primary crash to stop and remove all nodes from master and bind them to the slave after which all the client requests are forwarded to the slave.

\par They provide a nice component-based extensible design and show how multiple fault-tolerant mechanisms can be introduced into the computation graph. A strong feature of this approach is high availability of the ROS master, but it comes with the cost of adding an extra layer of communications between the nodes and the master.

\par Although this approach shows promise, the results for a ROS implementation, as reported by the authors themselves, were not satisfactory. Unfortunately, neither the source code nor the results of any experiments that were conducted with the proposed FTM in ROS are available for us to do a comparative study against our method.

\par Another approach by Jain et al\cite{jain2017dmtcp} uses DMTCP checkpointing \cite{Ansel2009} software to deal with the single point of failure of ROS master. DMTCP checkpointing is a multi threaded program which works by taking a snapshot of the application state which can be used later to restart the application from checkpoint. It is mainly designed for long running applications on a large cluster, where if the program crashes at the last minute, it needs to be started from the beginning.

In the DMTCP checkpointing \cite{Ansel2009} method an application snapshot which consists of the user space, the libraries, the processor state etc., are all saved. This is useful in case of long running applications where the entire application state needs to be recovered to continue processing. In ROS, the master acts as a service registry, where recovering just the meta-data is sufficient.

\par Also, DMTCP works by spawning a DMTCP coordinator thread and one checkpointing thread per node  which adds additional load on the processor. It saves one checkpointing image file per node. The checkpointing files are comparatively large in size, about 16MB for a simple string publisher node. This is due to the fact that the checkpoint contains the snapshot of the application state including the libraries, memory contents etc. This limits the scalability of this method. In addition, since, DMTCP  is a separate application from ROS it needs to checkpoint  periodically, each time suspending the nodes, saving the checkpoint and only then resuming the operation. This would entail a severe limitation for usage in robotics applications in highly dynamic environments. 
\par In the adaptive fault-tolerance solution, a replica master is maintained which replaces the primary in case of a failure. While this provides high availability of the ROS master, it adds the overhead of dynamic binding (unregistering/re-registering) all the  nodes to the replica and vice versa for the case when the master is recovered. It also requires a suspension of all the nodes when changing the bindings to the new master.
\par For DMTCP, we note that this case is quite different from that of ROS, where the master acts as a service registry. This implies that ROS can be restarted at any point without running the entire program from the beginning. Saving its meta-data is sufficient for recovery  and for re-establishing connections between nodes. No experimental evaluation of using DMTCP with ROS is provided in the paper to understand its applicability.

\begin{figure}[h]
\includegraphics[width=\textwidth]{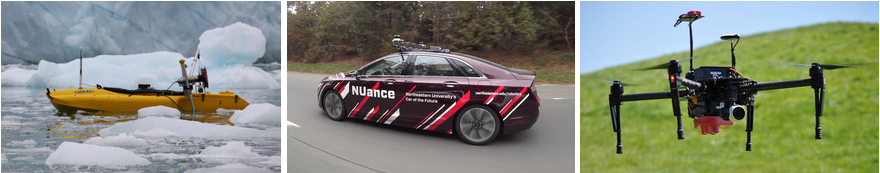}
\centering
\caption{ROS is the dominant software framework \cite{Quigley2009} used in variety of robotic platforms including aerial systems, autonomous driving and underwater \& surface vehicles. The requirement for a ROS master can be a critical point of failure, albeit with low probability. We propose a very lightweight method that addresses this issue.}
\end{figure}

\par In contrast to these methods, our work presents a simple and light-weight fault-tolerance mechanism as an added functionality within the ROS framework. We solve the problem of fault tolerance by maintaining the state of ROS master and its nodes via a logging mechanism. This not only helps us to store the meta information of the system but also helps us to recover the system state after a failure. It is simple as it is just a feature addition to ROS. It is light-weight in so far as we are only recording the meta-data and when there is a change to the configuration as opposed to periodic checkpointing. We believe our “ROS Rescue” is a very useful feature to have in ROS for carrying out real-world robotics missions/applications safely and without interruptions. A strong fundamental requirement for our work was to enable ROS master recovery in a fraction of a second, a number determined by the tolerance of our robotic systems in real world applications. The main contributions of this chapter include software for ROS master with fault tolerance, and a detailed step by step tutorial on how to obtain, build and use ROS Rescue into standard applications while meeting our fundamental requirement. 

\par In the coming sections of the chapter we discuss the following topics as they relate to our problem and its implementation
\begin{itemize} 
\item Section 2 covers an overview of the structure and various components of ROS.
\item Section 3 describes the impact of master failure and need  for fault tolerance in ROS.
\item Section 4 covers fault tolerance using our package which we call ROS Rescue.
\item Section 5  covers the implementation details of our fault tolerance technique and a step by step guide to install and use ROS rescue from an  applications standpoint.
\end{itemize}

\section{ROS Overview \& Architecture}
An overview of ROS framework is shown in the Figure 2 below. The main technical components include a ROS master, a parameter server and ROS nodes. Each of these components implement an XMLRPC server for communication. 
\begin{figure}[h]
\includegraphics[width=\textwidth]{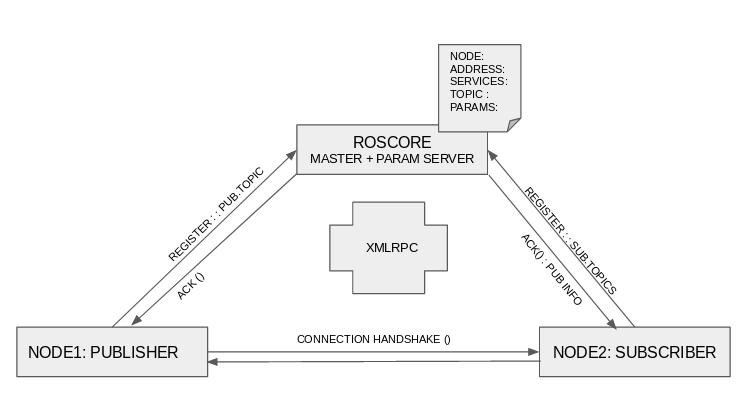}
\centering
\caption{The ROS software architecture in current state}
\end{figure}
\par As a peer to peer topology ROS requires some sort of look up service like a DNS server so that processes can find each other. The ROS master facilitates this and acts as the registration and lookup service which maintains a registration API for nodes to register with it as publishers, subscribers or service providers. The ROS master has a URI stored in the environment variable ROS\_MASTER\_URI on which the  XMLRPC server is running.
\par Every node is essentially an XMLRPC server and has a URI. This server is used to receive call backs from the master and negotiate connections with other nodes via a slave API. Nodes communicate with each other asynchronously using topics via a publish/subscribe mechanism, and synchronously via service calls.
\par A subscriber node connects with the master on startup. The master saves the information about the subscriptions and sends the publisher information which contains the subscribed topic names and the list of URIs of the nodes that publish those topics. The master also sends the updated publisher list to the subscriber in case of a change. After receiving publisher updates, the subscriber node will connect to any new publishers by requesting a topic connection using the publisher’s XMLRPC server. The nodes then communicate with each other on an agreed protocol (TCPROS or UDPROS).
\par Similarly, a publisher node establishes connection with the master first which saves the information about the advertised topics  and relays it to the subscribers. Publisher then receives calls from the nodes looking for topic connections for subscription. The topics are transported using libraries called TCPROS or UDPROS depending on the protocol.
\par In case of services, a node providing that service registers with the master. A client node requesting that service contacts the master and looks up the information. It then makes a serialized request call and gets the response from the service. In case of services there is no callback on the master which informs the clients about any changes. Thus, the clients poll and wait for the service to appear.
\par In addition to the above mentioned components there is a parameter server, which is a shared global dictionary consisting of the configuration parameters. The parameter server runs inside the ROS master and nodes can access and modify these parameters via remote procedure calls to the ROS master.

\par Thus, we can see that the master is an important entity which facilitates the communication between various nodes and stores the configuration state of  the runtime graph of a ROS network. Hence, it is crucial to keep the master up and running for system integrity.

\section{Fault tolerance in ROS}
The ROS architecture described above is characterized by single point of failure with an inadvertent loss of the ROS master. The master is the central server which keeps the metadata about all the nodes. The metadata is the information regarding the publisher nodes and subscriber nodes, their URIs, services and configuration parameters in parameter server. Every node on startup communicates with the master and registers itself. When the master fails this metadata is lost and the system is left with no way of connecting nodes. As it is a peer to peer system, nodes which are already connected still communicate. But, the problem arises when a new node comes up or goes down or an existing node makes a service call; the nodes need to contact the master for metadata information in these cases. Without recovering the lost metadata nodes are forced to restart even after master comes back alive and the system is reinitialized.

\par Consider a scenario with a publisher and a subscriber node. As explained in the previous section, after they register with master they communicate with each other directly. Now if the master crashes the nodes still talk to each other over the topic. However they cannot make any new connections or publish/subscribe new topics. When the master recovers, it starts afresh and does not have the knowledge of the current system state. This leaves the nodes hanging and isolated from any other nodes that may come up after reinitialization.
\par At present there is no logging or checkpointing mechanism in ROS to record the master metadata so that we can recover the last uncorrupted state. 

\subsection{Importance of Fault Tolerance in robotics}
Fault tolerance is a crucial property to have when implementing real time applications, especially in the field of robotics where the systems are highly prone to failures. There can be physical failures due to long deployments of missions in challenging environments such as at disaster areas or construction sites and in harsh conditions such as is the case for robotics underwater,  under-ice or in polar regions. There are additional issues such as battery limitations and power consumption, software failures including those associated with memory or processing overload. Moreover, in robotics the impact of a failure can have serious consequences including loss of the robot, property and lives. Due to the active and dynamic nature of these systems, the time to recovery is a critical factor. However, the time to recovery from failure depends on the type of robot, the underlying computing hardware and reaction time of sensors and/or actuators, all of which vary with the type of robot and its application area. 

\par The main goal of this chapter is to provide an implementation of a fault tolerant mechanism in ROS, which is the most widely used software framework \cite{Quigley2009} for building robotics applications both in academia and industry in order to facilitate scalable and reliable robotic systems. In the following sections, we lay out our method including the logging and recovery solution, and its implementation.

\section{ROS Rescue}
We have designed, developed and tested our software which we call ROS Rescue, that deals with the two related problems of system failure detection and recovery. The fundamental idea is to equip the ROS master with a logging mechanism to save its state to persistent memory whenever there is a change to its metadata. When the master recovers from failure it can return to the previous working state by reading the log. The log contains the complete metadata information about various nodes including URIs, port numbers, published/subscribed topics and services, and the parameters in the parameter server. After recovering the working state, the master can resume its operation.

\par We  developed our approach to be as simple and light weight as possible and at the same time keep it suitable for robotics applications.

Our solution is a simple feature addition to ROS master as opposed to running another multi-threaded application or having interceptor nodes and a replica master. We save only the meta-data as a log file as opposed to the application snapshot consuming less space. We log if and when a change occurs to the ROS master instead of periodically saving a checkpoint. And, we are not suspending the nodes when we are writing to persistent disk, in fact this can be parallely done using a writing thread for complete separation. Finally, we can recover the configuration by just restarting the master while all the other nodes are active unlike the checkpoint recovery using DMTCP.

\par The architectural changes to the original ROS framework can be visually depicted as shown below in Figure 3. The important aspects of the fault tolerance mechanism are:

\begin{figure}[h]
\includegraphics[width=\textwidth]{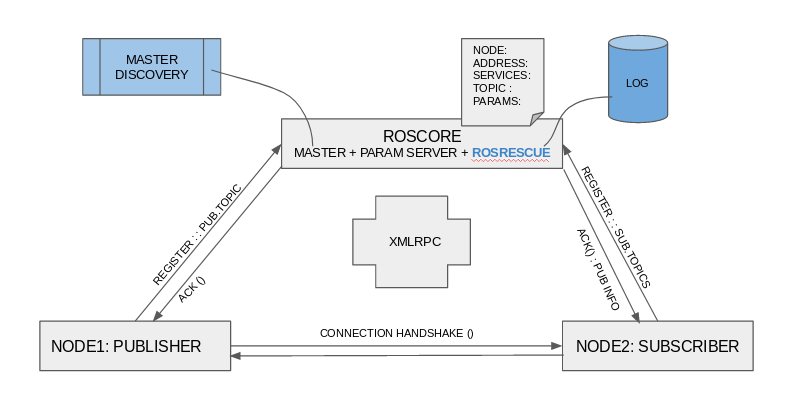}
\centering
\caption{ ROS Software architecture with fault-tolerance. In comparison to Figure 2 note the addition of ROSRESCUE inside ROSCORE along with an associated logging mechanism}
\end{figure}

\begin{enumerate}
\item Failure detection by continuous monitoring of the ROS master.
\item Detecting changes in the network configuration and continuous update of meta information to persistent memory.
\item Recovery of the most recent state after master crash and updates to the state to reflect the network configuration in a timely manner.
\item The code simply provides added functionality, while all the key components, their functions, and APIs, otherwise remain the same within ROS. There is no overhead associated with external libraries.
\item The software is lightweight and is a simple implementation when compared to  previously proposed solutions.
\end{enumerate}

\subsection{ Implementation Details}
The architecture of ROS is very similar to that of the google file system (GFS) \cite{Ghemawat2003} where there is a single master which saves the meta information and chunk servers which host the data. When needed the clients and chunk-servers establish connections after contacting master for look up information. The implementation of logging and recovery mechanism is inspired from GFS , but we record the latest state of the master as opposed to an operational log which needs to be replayed for recovery. In case of GFS an operational log is maintained not just for metadata but also to record the logical timeline of the concurrent operations on the data and maintain version information. In case of ROS, this is not necessary as it is not a database or filesystem but a framework of processes which only needs the information about about process communication. The fault tolerance is implemented in two stages

\subsubsection{Failure detection} A master discovery node is implemented which keeps track of the availability of the master. It starts up with the master and detects master failure using a pull design by periodically polling the master node. Finding the master state is done by repeatedly making a XMLRPC function call to the master server. When it detects that the master has not responded for a specified number of polls it declares that the master has failed and issues command to restart the master.
\subsubsection{Recovery}  master logging and recovery is implemented inside the ROS Rescue process which exists within the master and can be activated by passing a command line argument to the rosmaster startup script. Once this feature is enabled ROS Rescue repeatedly logs the metadata information of master onto persistent disk whenever it changes.  This change can happen when a new node registers/unregisters itself with the master, a node crashes or stops, when a new topic is published or subscribed to, or changes are made to parameter server. When the master recovers from failure it restores its last seen state by reading the log. It then checks to see if there are any changes to the recovered state. For example, some nodes might have crashed while the master is down but their information will be present in the log. Hence, master goes through each node and makes sure that they exist and their metadata remains consistent. If there are any changes it updates its state and re-establishes connections with the nodes that are running.

\begin{figure}[h]
\includegraphics[width=\textwidth]{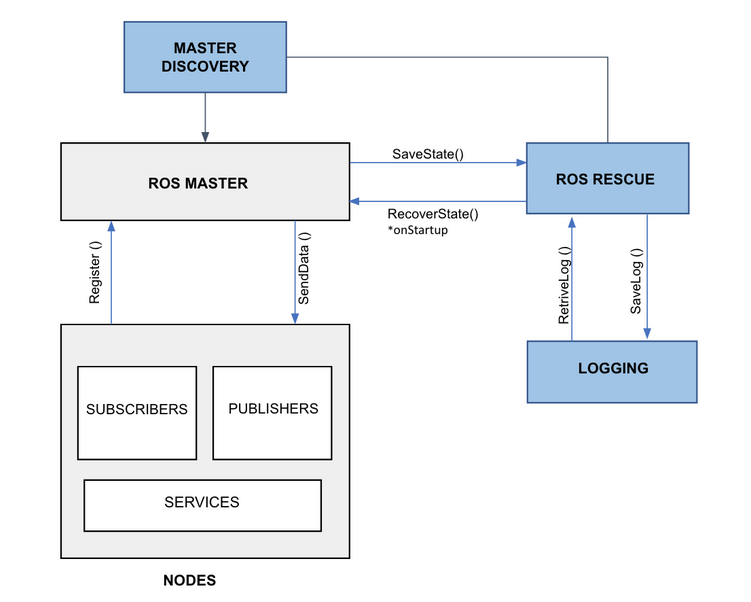}
\centering
\caption{ Process flow diagram of the Fault tolerance mechanism}
\end{figure}

\subsection{Logging meta-data}
As mentioned in the previous section the state of the ROS runtime graph is written to a log whenever there is a change.  ROS master maintains the information about publishers, subscribers and the available services in the form of dictionaries.  The structure of the log that is written to disk is maintained in a manner similar to the way it is stored in the ROS master for easier recovery. The yaml format is a natural choice to save the metadata of ROS master and the written log has the  structure shown in Figure 5 and illustrated for a real process with all its complexity in Figure 7. 
\begin{figure}[h]
\includegraphics[width=7cm]{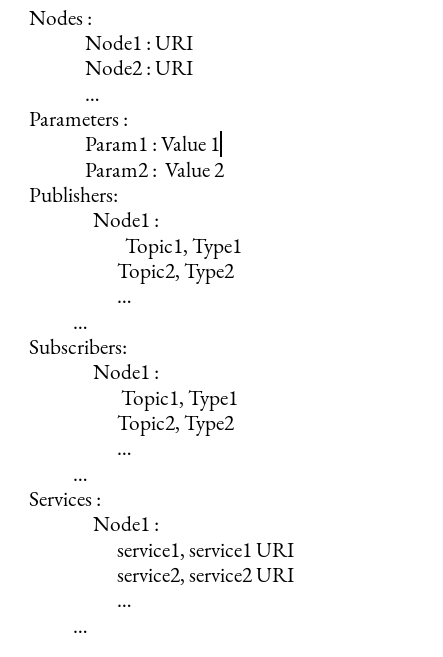}
\centering
\caption{ Log file structure. Our log file is in YAML format and contains all the metadata associated with the ROS master. In comparison to this simple illustration, we have also included the partial contents of an actual log file in Figure 7.}
\end{figure}

\subsection{Testing}
The following cases are considered for testing the implemented fault tolerance and recovery mechanism.\\

Case 0 -Master failure and no configuration changes:
Fault tolerance mechanism in this case is demonstrated by spawning three nodes - ROS master, a publisher node advertising a topic and a subscriber listening to the same topic. We can then abruptly crash the master and upon restart the master should be able to get back it’s previous state and ready to connect new nodes. We can check this by:
\begin{enumerate}
    \item Inspecting the meta data and the configuration parameters
    \item Spawning a new subscriber to the topic advertised by the publisher before the crash and checking on recovery, to ensure that the subscription was successful.
\end{enumerate}

Case 1 - Master failure and node configuration changes:
This case is demonstrated similar to case 0, but now we crash one of the nodes while the master is down. Upon restart, the master should not only recover it last seen state, but also update the latest node configuration changes. This can be tested in a way similar to the last case.

Case 2 - Master failure during the registration process: 
By default ROS deals with this issue as typically a node keeps trying periodically to see if the master is available during the registration process. However, there can be a situation where the master crashes after it unregisters a node with itself, but before it can let other nodes know about the change. This case is currently not being handled in the code although we are working on the implementation. Our method would go through the topics published by the nodes and notify subscribers that are alive but are not privy to this information as observed in there registration data.

Case  3 - Master failure while writing to persistent disk:
If the master fails while it is saving the state it is important to save either the entire meta-data or none at all to avoid inconsistent state recovery. We have implemented a two stage commit for saving the system state. Whenever the master saves the state, it writes to a temporary checkpoint file which will be later renamed as the latest checkpoint upon successful write. That way, if the master dies/crashes while writing the state it will be in the temporary file and we can still recover the last fully saved state from the old checkpoint.

\par All the above mentioned cases have been tested  using ROS Rescue and in each case we were able to recover the meta-data of the ROS runtime graph. In order to evaluate the effectiveness of our fault-tolerant approach we also performed quantitative analysis in terms of time taken for master recovery for a varying number of nodes and on different robotic systems. The first set of experiments was conducted on a Ubuntu 16.04, amd64 desktop where the ROS master was run with and without the rescue option. For each test, we spawned a set of n nodes (both publishers and subscribers), killed the master and brought it back up. We recorded the time taken for recovering the saved state from the log file. This test was performed multiple times for each value and the results averaged. The final averaged results are tabulated in Table \ref{tab1}. We can see that time to recovery changes linearly with the number of nodes. We also compare these results with a vanilla version of ROS (that is ROS without ROS Rescue compiled in).  

We have successfully tested and have actively been using ROS rescue on multiple robotic platforms including the turtlebot3 \cite{turtlebot}, DJI matrice M100\cite{dji} and a Lincoln MKZ autonomous car equipped with the drive-by-wire ADAS kit from Dataspeed Inc\cite{dataspeed}. These robotic platforms use the Raspberry PI, Nvidia Tx2 and an Intel Xeon processor (running Ubuntu) respectively. We have tabulated the results of these experiments in Table \ref{tab2}. For each robot we include the average number of nodes while running, the average time taken for ROS master recovery and the start time for ROS with and without rescue option. We can see that the recovery time is different on different platforms, and unlike Table \ref{tab2}, is not linear. We note that this is to be expected as in these cases, the computational burden on the differing underlying computer hardware is different. However, in each case the time to recovery is well under a second which was our required figure of merit based on the dynamic constraints for these robotic systems. 

\begin{table}[]
\centering
\caption{ROS Rescue experimental results showing the total time taken to start the master and time taken  to recover the master in ROS with and without ROS Rescue. }
\label{tab1}
\resizebox{8cm}{!}{%
\begin{tabular}{|l|l|l|}
\hline
\textbf{} & \textbf{\begin{tabular}[c]{@{}l@{}}Time to \\ recovery\end{tabular}} & \textbf{\begin{tabular}[c]{@{}l@{}}Total time to\\ start master\end{tabular}} \\ \hline
Vanilla ROS & N/A & 0.0579s \\ \hline
ROS Rescue 1 node & 0.00512s & 0.0638s \\ \hline
ROS Rescue 5 nodes & 0.0593s & 0.1173s \\ \hline
ROS Rescue 10 nodes & 0.112s & 0.175s \\ \hline
ROS Rescue 20 nodes & 0.218s & 0.276s \\ \hline
ROS Rescue 40 nodes & 0.436s & 0.499s \\ \hline
ROS Rescue 80 nodes & 0.9s & 0.97s \\ \hline
\end{tabular}%
}
\end{table}

\begin{table}[]
\centering
\caption{Experimental results on real robotic platforms - turtlebot3, DJI matrice M100 drone and autonomous car showing the average time taken to recover the master state and total time to start the master with and without ROS Rescue for a typical application. As explained in the text the results are not linear with the number of nodes due to different underlying computer hardware. However, full recovery is well within the sub second figure of merit. }
\label{tab2}
\resizebox{\textwidth}{!}{%
\begin{tabular}{|l|l|l|l|}
\hline
 & Vanilla ROS & \begin{tabular}[c]{@{}l@{}}Avg time to \\ start master\end{tabular} & \begin{tabular}[c]{@{}l@{}}Avg time to \\ recover\end{tabular} \\ \hline
Turtlebot3 (7$\sim$10 nodes) & 0.23s & 0.62s & 0.36s \\ \hline
Drone (15-20 nodes) & 0.12s & 0.45s & 0.35s \\ \hline
Autonomous Car (20-25 nodes) & 0.057s & 0.33s & 0.27s \\ \hline
\end{tabular}%
}

\end{table}

\section{Step-by-step guide to ROS Rescue}
This section provides the necessary guidelines for building, testing and implementing ROS Rescue in custom ROS applications.

\subsection{System setup}
Ros Rescue is included as an additional functionality inside the rosmaster package for light-weight implementation. We have implemented the rescue feature in the bare bones source code of ROS which is available on github at the repository shown below.

\url{https://github.com/PushyamiKaveti/fault-tolerant-ros-master.git}\\

We need to prepare the system to execute  roscore with rescue feature enabled , as a first step we make sure to have the prerequisites. Open the terminal and execute the following commands which take cares of python dependencies, initialize the ROS dependencies and update the ROS dependency packages.\\

\begin{lstlisting}[language=sh , frame=single , basicstyle=\small\ttfamily, breaklines=true]

$ sudo apt-get install python-rosdep python-rosinstall-generator python-wstool python-rosinstall build-essential
$ sudo rosdep init
$ rosdep update

\end{lstlisting}

\par To use ROS Rescue the code needs to be built from source. Open terminal and navigate to the folder of your choice  and clone the git repository mentioned above \\

\begin{lstlisting}[language=sh , frame=single , basicstyle=\small\ttfamily, breaklines=true]

$ git clone https://github.com/PushyamiKaveti/fault-tolerant-ros-master.git
  
\end{lstlisting}

\par Now,  issue the following commands in the terminal. This will build the code locally in the current folder and run ROS with the fault tolerant feature enabled. \\

\begin{lstlisting}[language=sh , frame=single , basicstyle=\small\ttfamily, breaklines=true]

$ cd fault-tolerant-ros-master
$ ./src/catkin/bin/catkin_make_isolated --install -DCMAKE_BUILD_TYPE=Release 

\end{lstlisting}

\par Open the ros\_env\_setup.sh file present in the fault-tolerant-ros-master folder in a text editor. Find and replace all \textless your\_path\_folder\textgreater with the your local path. For example, if you have cloned our repository in home folder then the  \textless your\_path\_folder\textgreater will be replaced as  `` \textasciitilde/fault-tolerant-ros-master'' the absolute path of the file location.

\par Also, the environment setup file has the list of environment variables which are required for execution. Sourcing the file will keep the current shell aware of the environment variables. If you want to utilize the ROS for other users in the system, place the copy of the file in “/etc/profile.d/ros\_env\_setup.sh”. This assumes you are running ROS on a linux/unix platform. This will facilitate loading the file globally for all users in the system during login/shell spawning. \\

\begin{lstlisting}[language=sh , frame=single , basicstyle=\small\ttfamily]

$ source ros_env_setup.sh

\end{lstlisting}

\par The system setup instructions are needed to build ROS Rescue from source for immediate use. However, we plan to limit the ROS Rescue functionality only to rosmaster package and make a pull request to ROS's offical github repository. Once approved the Rescue functionality will be available in official ROS packages and one can skip to section 5.2 for running ROS with rescue option.

\subsection{Running ROS Rescue}
Once the environment is setup as outlined above we can run roscore with the rescue feature enabled. ROS master with our rescue option will function and operate as normal with the additional capabilities associated with logging and rescue. The only change is to initiate the roscore with ``--rescue'' option as shown below. Before running roscore it is advisable to ensure that the correct version of roscore is being called by issuing the ``which roscore'' command in the terminal.\\

\begin{lstlisting}[language=sh , frame=single , basicstyle=\small\ttfamily]

$ roscore --rescue

\end{lstlisting}

\par By default logs are saved in ``\textasciitilde/.ros/log/latest-chkpt.yaml'' with the system default’s umask value which enables everyone to read the logs. To keep the logs locked or to avoid manual intervention you can set the file permission to `655'. Owner can read and write, others can only read and execute by ``chmod 655  \textasciitilde/.ros/log/latest-chkpt.yaml''. This will help us keep the master state secure and stable.

\par If the ROS master is running with fault tolerance enabled you should see ``ROS Rescue enabled. Master is now fault tolerant!!'' printed to the console where roscore --rescue was executed as shown in the Figure 6. Another check that can be done is to make sure the master state is being saved to latest-ckhpt.yaml file mentioned above.

\begin{figure}[h]
\includegraphics[width=\textwidth]{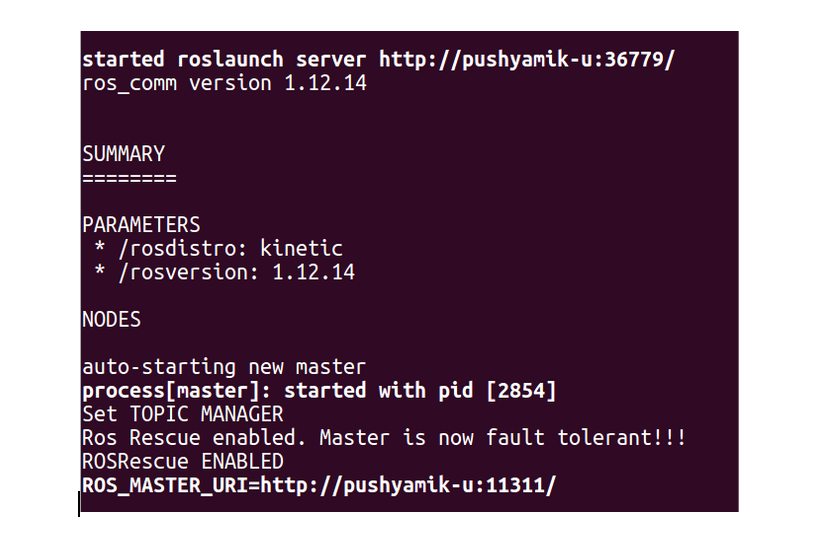}
\centering
\caption{Console output of roscore with fault tolerance enabled}
\end{figure}

ROS Rescue package also comes with a master discovery node to detect master failure.  Master monitoring is required for failure detection and master restart. This node is made a part of roscore.xml file alongside rosout node and is launched when roscore command is issued. We can also run it separately by issuing the following commands. \\

\begin{lstlisting}[language=sh , frame=single , basicstyle=\small\ttfamily]

$  cd src/ros_comm/rosmaster/scripts/
$ ./master_monitor --rescue

\end{lstlisting}

To make sure ROS Rescue logging is working correctly  we can run a simple publisher/subscribe test as shown below.\\

\begin{lstlisting}[language=sh , frame=single , basicstyle=\small\ttfamily]

$ cd  src/node_test/scripts/
$ python  talker.py
$ python listener.py

\end{lstlisting}

\par The  talker node publishes a dummy string message to a topic called ‘/chatter’. The listener node subscribes to that string message on topic “/chatter”.  If everything is working well the messages published by publisher in its console should relay on subscriber console.  The log file should now contain the information about the nodes, publishers, subscribers and services available. Figure. 7 below shows a small portion of a log file that was created with ROS Rescue  enabled when talker and listener nodes were running.

\begin{figure}[h!]
\includegraphics[width=\textwidth]{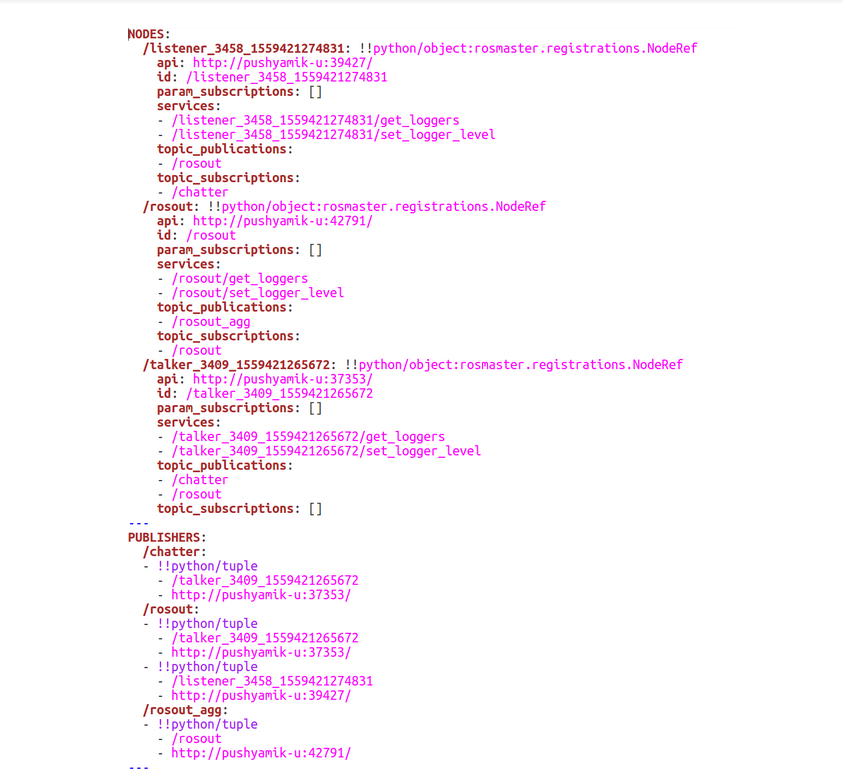}
\centering
\caption{Screenshot of an actual run showing the latest-chkpt.yaml created with fault tolerance enabled}
\end{figure}

\section{Conclusion and Future work}
As part of this project a failure detection and recovery mechanism has been implemented for Robot Operating System 1.0 (ROS). The implemented code has been tested successfully on custom publisher and subscriber nodes as well as on a real robot systems. We also conducted tests and used ROS Rescue on a turtlebot3, an autonomous car and on drones. The logging mechanism has implemented for publishers, subscribers, services and the parameter server. We have made our code and demo videos available on the project\rq s github repository for use by researchers and developers in the ROS community as they see fit. As mentioned earlier we plan to limit ROS Rescue functionality to rosmaster package and make a pull request to the official ROS repository for integration and porting across multiple versions. As an extension to this work we are in the process of conducting testing to cover cases of highly dynamic and scalable environments, including for a multi-robot swarm system.

\section{Authors Biographies}
\textbf{Pushyami Kaveti}  is now a Ph.D. student at the Khoury College of Computer Sciences at Northeastern University . She received her M.S. degree in Computer Science in 2014 at University of Florida , and B.Tech degree in Computer Science and Engineering from JNTU, India in 2011. Her research interests lie at the intersection of robotics, computer vision and machine learning. Her work at Field Robotics Laboratory focuses on robot perception and navigation in real world and dynamic environments.\\

\textbf{Hanumant Singh} is professor in the department of ECE at Northeastern University and the Director of the Center for Robotics there. He received his Ph.D. from the MIT/WHOI Joint Program in 1995 after which he worked on the Staff at WHOI until 2016 when he joined Northeastern. Prior to that, he graduated with dual degrees in Computer Science and Electrical and Computer Engineering from George Mason University. While at WHOI, his group designed and built the Seabed AUV, as well as the Jetyak Autonomous Surface Vehicle, dozens of which are in use for scientific and other purposes across the globe. His interests lie in the area of Marine, Land and Aerial robotics especially as they relate to sensing, imaging and navigation.

\bibliographystyle{splncs04}
\bibliography{references.bib}

\end{document}